\documentclass[conference,a4paper]{IEEEtran}
\IEEEoverridecommandlockouts
\pdfpagewidth=\paperwidth
\pdfpageheight=\paperheight
\usepackage{cite}
\usepackage{amsmath,amssymb,amsfonts}
\usepackage{graphicx}
\usepackage{textcomp}
\usepackage{xcolor}
\usepackage{booktabs}
\usepackage{array}
\usepackage{url}
\usepackage{tikz}
\usetikzlibrary{arrows.meta, positioning, fit, backgrounds, calc}

\makeatletter
\newcommand{\twocolumnfootnotefullwidth}[1]{%
  \begingroup
  \renewcommand{\thefootnote}{}%
  \footnotetext{%
    \noindent\hspace*{-1em}\rule{0.3\linewidth}{0.4pt}\\[0.2em]
    \vspace*{-1em}
    \noindent\footnotesize #1
  }
  \endgroup
}
\makeatother

\begin{document}

\title{LEXIC: Lightweight On-Device Decoding of Reading Comprehension from Eye Movements}

\author{%
\makebox[\textwidth][c]{%
\begin{tabular}{@{}c@{\hspace{1.0em}}c@{\hspace{1.0em}}c@{\hspace{1.0em}}c@{}}
Sumin Lee & Kyeonghun Kim & Subeen Lee & Jiwon Yang \\
\textit{Seoul National University} & \textit{OUTTA} & \textit{Seoul National University} & \textit{Seoul National University} \\
{\fontsize{10}{11}\selectfont cirtuare@snu.ac.kr} &
{\fontsize{10}{11}\selectfont kyeonghun.kim@outta.ai} &
{\fontsize{10}{11}\selectfont 5584sb@snu.ac.kr} &
{\fontsize{10}{11}\selectfont jwyang29@snu.ac.kr}
\end{tabular}}%
\\[1.3ex]
\makebox[\textwidth][c]{%
\begin{tabular}{@{}c@{\hspace{1.0em}}c@{\hspace{1.0em}}c@{\hspace{1.0em}}c@{}}
Hyunsu Go & Eunseob Choi & Ken Ying-Kai Liao & Nam-Joon Kim\textsuperscript{\dag} \\
\textit{Seoul National University} & \textit{GIST} & \textit{NVIDIA} & \textit{Seoul National University} \\ 
{\fontsize{10}{11}\selectfont hsmail02@snu.ac.kr} &
{\fontsize{10}{11}\selectfont eunseobchoi@gm.gist.ac.kr} &
{\fontsize{10}{11}\selectfont kenyingkail@nvidia.com} &
{\fontsize{10}{11}\selectfont knj01@snu.ac.kr}
\end{tabular}}%
}

\maketitle
\twocolumnfootnotefullwidth{{\dag} Corresponding author}

\begin{abstract}
Predicting comprehension from eye movements could support adaptive reading interfaces.
We present LEXIC, a compact recurrent model that predicts response correctness from fixation sequences, word frequency, and character length.
It has 41.6K parameters and requires no language-model inference.
Mean area under the receiver operating characteristic curve (AUROC) reaches 0.529 for Unseen Text and 0.554 for Unseen Reader on OneStop.
Matched comparisons with AhnCNN show gains from both encoder redesign and lexical augmentation in these settings.
Separate training and evaluation on SB-SAT also yield higher mean AUROC than AhnCNN. LEXIC occupies only 166 KB of model weights and requires 1.4 ms per trial on a single CPU core, supporting lightweight on-device inference
\end{abstract}

\begin{IEEEkeywords}
eye tracking, reading comprehension, on-device inference, lightweight neural networks, consumer electronics
\end{IEEEkeywords}

\section{Introduction}

Eye movements during reading reflect lexical processing and comprehension~\cite{just1980theory,rayner1998eye}. We study whether a reader's fixation sequence for a passage can predict the correctness of a subsequent comprehension response. Such predictions could support adaptive reading interfaces~\cite{shubi2025eyebench}.

Existing approaches include predictors based on aggregated eye-tracking measures, neural fixation-sequence models, and models incorporating linguistic features or pretrained language models~\cite{meziere2023using,ahn2020rcn,reich2022beyelstm,shubi2024finegrained}. EyeBench provides shared data splits for evaluating generalization to new readers and texts~\cite{shubi2025eyebench}. We use these splits to assess how encoder design and the addition of word frequency and character length affect prediction relative to a neural gaze-only baseline.

We make three contributions:
\begin{itemize}
  \item LEXIC, a 41.6K-parameter bidirectional gated recurrent unit (GRU) model that combines variable-length fixation sequences with word frequency and character length.
  \item An analysis of the predictive contributions of encoder design and lexical features, including comparisons with GPT-2 surprisal.
  \item Evaluation on OneStop and SB-SAT, together with an on-device efficiency analysis covering model size, operation count, and CPU inference latency.
\end{itemize}

\section{Related Work}

\textbf{Comprehension prediction from gaze.} Eye movements reflect attention, word recognition, and comprehension~\cite{rayner1998eye, kliegl2004}. Earlier studies used reading time, fixation duration, regressions, and skipping rates with classical classifiers~\cite{meziere2023using, makowski2018discriminative}. Neural models learn from fixation sequences~\cite{ahn2020rcn}, add linguistic features~\cite{reich2022beyelstm}, or combine gaze with pretrained text encoders~\cite{yang2023plmas, shubi2024finegrained}. EyeBench provides a common evaluation protocol across OneStop, SB-SAT, and MECO, including tests on unseen texts and readers~\cite{shubi2025eyebench, berzak2025onestop, ahn2020rcn, kuperman2021meco}. Related work predicts the reader's task~\cite{shubi2025decoding} or uses gaze to support text classification~\cite{barrett2018sequence}.

\textbf{Lexical features.} Word frequency, length, and surprisal are established predictors of fixation behavior~\cite{hale2001probabilistic, smith2013effect, kliegl2004}. These features can be computed once per text and reused across readers. Performance remains less conclusive in the Unseen Both setting, where the smaller test set produces higher variance and the improvement over AhnCNN is not statistically significant. LEXIC uses frequency and length from lookup tables. We evaluate surprisal separately to measure the benefit of adding a language model.

\section{Method}

\begin{figure*}[t]
\centering
\includegraphics[width=\textwidth]{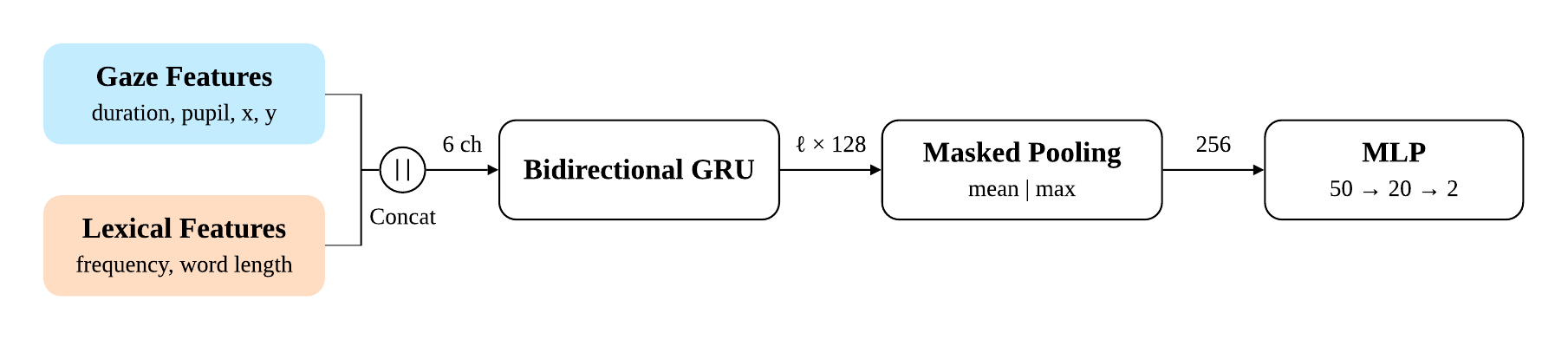}
\caption{LEXIC combines four gaze features with word frequency and length at each fixation. A bidirectional GRU processes the true sequence length $\ell$. Masked mean and max pooling produce a fixed size representation for the MLP, which predicts whether the comprehension response is correct.}

\label{fig:arch}
\end{figure*}

LEXIC combines the gaze encoder and two lexical features described below (Fig.~\ref{fig:arch}).

\subsection{Gaze Encoder}
Each trial contains a sequence of fixations with four features: duration, pupil size, and position $(x,y)$. EyeBench pads sequences to $815$ steps, while the mean sequence contains about $113$ fixations. LEXIC processes each sequence at its true length using a bidirectional GRU with $64$ hidden units per direction. Masked mean and max pooling exclude padded steps and produce a $256$-dimensional representation. The classifier uses layers of size $50 \to 20 \to 2$ with dropout $0.3$, matching the baseline head. Removing the lexical features gives the ablation model, LEXIC without lexical features, with $40{,}792$ parameters.

\subsection{Lexical Features}
We append corpus frequency from \texttt{wordfreq}~\cite{speer2018wordfreq} and character length to each fixation, aligned by the fixated word. This increases the input from four to six features and adds $768$ parameters. Values are computed once per text, scaled using training data, and set to zero where alignment is unavailable. Both features use lookup tables. The complete model has $41{,}560$ parameters. We also evaluate LEXIC with frequency only, which omits length, and LEXIC with surprisal, which adds GPT-2 surprisal~\cite{radford2019gpt2} computed offline for each text.

\subsection{Inference Cost}
LEXIC requires $3.1$ million multiply-accumulate operations (MMACs) at the mean sequence length and $21.9$ MMACs at $815$ fixations. Lexical features require one lookup per fixated word. Section~\ref{sec:cost} reports measured inference latency.

\section{Experimental Setup}

We follow the EyeBench v1.0 protocol~\cite{shubi2025eyebench}. \textbf{OneStop}~\cite{berzak2025onestop} contains $9{,}718$ trials from $180$ participants and $324$ passages. We use the ordinary reading condition, with no question preview and only the first reading. The target is a correct or incorrect comprehension response; $81.2\%$ of responses are correct. Evaluation uses ten cross-validation folds in three settings: \emph{Unseen Text}, \emph{Unseen Reader}, and \emph{Unseen Both}. \textbf{SB-SAT}~\cite{ahn2020rcn} uses four folds with the same settings.

All models use five seeds per fold, AdamW with a learning rate of $3\times10^{-4}$, gradient clipping of $1.0$, and batch size $16$. Training runs for up to $1000$ epochs, stopping after $50$ epochs without validation improvement. We select the checkpoint with the lowest validation cross-entropy. Features are normalized with a robust scaler fitted on the training fold.

We report AUROC averaged across five independent runs per fold, then report the mean and standard deviation across folds. Aggregate scores pool test predictions from the three settings within each seed and fold before computing AUROC, then use the same averaging. Predictions are not ensembled. On OneStop, we also report paired fold differences, Wilcoxon signed-rank $p$-values~\cite{wilcoxon1945}, bootstrap $95\%$ confidence intervals, and the number of folds with positive differences. These statistics describe comparisons across folds that share training data~\cite{demsar2006statistical}. Unseen Both has about $97$ test trials per fold; the other settings have about $875$.

\textbf{Baselines.} AhnCNN and AhnRNN~\cite{ahn2020rcn} are retrained from the released EyeBench implementation using the same folds, seeds, and checkpoint selection. AhnCNN's pooled AUROC is within $0.1$ percentage points of the published EyeBench result. For the ablation, we also provide AhnCNN with the same frequency and length features (AhnCNN+F+L). BEyeLSTM and PLM results are published pooled AUROCs from EyeBench.

\section{Results}

\subsection{OneStop: Main Results}

\begin{table}[t]
\caption{OneStop test AUROC (mean $\pm$ std across ten folds, averaging five seeds within each fold). LEXIC includes frequency and length (F+L). Reference rows are published pooled AUROCs~\cite{shubi2025eyebench}.}

\label{tab:main}
\centering
\footnotesize
\renewcommand{\arraystretch}{1.05}
\setlength{\tabcolsep}{2.2pt}
\begin{tabular}{@{}lccc@{}}
\toprule
Model & Unseen Text & Unseen Reader & Unseen Both \\
\midrule
AhnCNN (retrained) & $0.496 \pm .016$ & $0.496 \pm .015$ & $0.510 \pm .047$ \\
AhnRNN (retrained) & $0.500 \pm .000$ & $0.500 \pm .000$ & $0.500 \pm .000$ \\
\textbf{LEXIC} & $0.529 \pm .028$ & $0.554 \pm .028$ & $0.520 \pm .071$ \\
\midrule
\multicolumn{4}{l}{\textit{Ablation}} \\
AhnCNN + F+L & $0.512 \pm .021$ & $0.514 \pm .027$ & $0.528 \pm .043$ \\
\shortstack[l]{LEXIC without\\lexical features} & $0.520 \pm .022$ & $0.521 \pm .022$ & $0.481 \pm .066$ \\
\shortstack[l]{LEXIC with\\frequency only} & $0.536 \pm .039$ & $0.548 \pm .029$ & $0.519 \pm .074$ \\
LEXIC with surprisal & $0.527 \pm .027$ & $0.560 \pm .027$ & $0.515 \pm .061$ \\
\midrule
BEyeLSTM (published) & \multicolumn{3}{c}{$0.525$ (pooled)} \\
Best PLM (published) & \multicolumn{3}{c}{$0.56$ -- $0.63$ (pooled)} \\

\bottomrule
\end{tabular}
\end{table}

LEXIC reaches $0.529$ AUROC on Unseen Text and $0.554$ on Unseen Reader (Table~\ref{tab:main}). It improves over retrained AhnCNN by $3.37$ and $5.78$ percentage points, respectively, with positive differences in nine of ten folds for both settings (Table~\ref{tab:paired}). The difference on Unseen Both is $0.96$ points ($p{=}.770$). Pooled AUROC is $0.497$ for AhnCNN and $0.542$ for LEXIC.

\begin{table}[t]
\caption{Paired within-fold deltas on OneStop. Each delta subtracts the reference model named below. U.\ Text / U.\ Reader / U.\ Both = the three regimes; sign = folds with positive delta out of ten; CI = percentile bootstrap $95\%$ interval, resampling the ten fold-level deltas ($10{,}000$ draws, random seed $0$ for each comparison).}
\label{tab:paired}
\centering
\footnotesize
\renewcommand{\arraystretch}{1.05}
\setlength{\tabcolsep}{2.2pt}
\begin{tabular}{@{}llrccc@{}}
\toprule
Model & Regime & $\Delta$ (pp) & sign & $p$ & CI (pp) \\
\midrule
\multicolumn{6}{l}{\textit{Reference: AhnCNN}} \\
\textbf{LEXIC}   & U.\ Text   & $+3.37$ & $9/10$ & $.006$ & $[+1.8, +5.2]$ \\
                 & U.\ Reader & $+5.78$ & $9/10$ & $.004$ & $[+3.7, +7.7]$ \\
                 & U.\ Both   & $+0.96$ & $5/10$ & $.770$ & $[-4.3, +6.2]$ \\
\midrule
LEXIC without    & U.\ Text   & $+2.44$ & $8/10$ & $.020$ & $[+1.0, +3.9]$ \\
lexical features & U.\ Reader & $+2.50$ & $7/10$ & $.027$ & $[+0.5, +4.3]$ \\
                 & U.\ Both   & $-2.96$ & $3/10$ & $.193$ & $[-6.7, +1.4]$ \\
\midrule
\multicolumn{6}{l}{\textit{Reference: AhnCNN + F+L}} \\
\textbf{LEXIC}   & U.\ Text   & $+1.79$ & $8/10$ & $.020$ & $[+0.6, +3.0]$ \\
                 & U.\ Reader & $+4.05$ & $9/10$ & $.004$ & $[+2.1, +6.1]$ \\
                 & U.\ Both   & $-0.84$ & $4/10$ & $.557$ & $[-3.8, +2.6]$ \\
\bottomrule
\end{tabular}
\end{table}

\subsection{Ablation}

\textbf{Encoder and lexical features.} Without lexical features, LEXIC improves over AhnCNN by $2.44$/$2.50$ percentage points on Unseen Text/Unseen Reader (Table~\ref{tab:paired}). Adding F+L to AhnCNN improves its AUROC by $1.58$/$1.73$ points ($p{=}.027$ in both settings). With F+L provided to both models, LEXIC remains higher by $1.79$ points on Unseen Text ($8/10$ folds, $p{=}.020$) and $4.05$ on Unseen Reader ($9/10$, $p{=}.004$), and is $0.84$ points lower on Unseen Both ($p{=}.557$).

\textbf{Choice of lexical features.} On Unseen Reader, frequency alone improves over LEXIC without lexical features by $2.71$ points ($p{=}.084$); frequency and length improve it by $3.28$ ($p{=}.027$). Adding surprisal to the complete LEXIC model changes AUROC by $+0.59$ points on Unseen Reader ($p{=}.084$) and $-0.22$ on Unseen Text ($p{=}.275$). The frequency only, complete, and surprisal variants have similar pooled AUROC ($0.543$, $0.542$, and $0.544$). LEXIC uses frequency and length, which require no language model computation.

\subsection{SB-SAT Results}

\begin{table}[t]
\caption{SB-SAT test AUROC of a single model (mean $\pm$ std, four folds, five seeds).}
\label{tab:sbsat}
\centering
\footnotesize
\renewcommand{\arraystretch}{1.05}
\setlength{\tabcolsep}{2.2pt}
\begin{tabular}{@{}lccc@{}}
\toprule
Model & Unseen Text & Unseen Reader & Unseen Both \\
\midrule
AhnCNN & $0.500 \pm .015$ & $0.500 \pm .033$ & $0.461 \pm .040$ \\
\shortstack[l]{LEXIC without\\lexical features} & $0.543 \pm .013$ & $0.504 \pm .022$ & $0.502 \pm .022$ \\
\textbf{LEXIC} & $0.546 \pm .013$ & $0.519 \pm .016$ & $0.517 \pm .009$ \\
\bottomrule
\end{tabular}
\end{table}

On SB-SAT (Table~\ref{tab:sbsat}), LEXIC improves over AhnCNN by $4.6$ percentage points on Unseen Text and $1.9$ on Unseen Reader. Without lexical features, the Unseen Text improvement is $4.3$ points. Both configurations improve on Unseen Text in all four folds.

\subsection{Model Inference Cost}
\label{sec:cost}

\begin{table}[t]
\caption{Model inference cost on an Apple M-series CPU (one thread, batch size 1, FP32). Latency is the median of seven run medians, each from 200 timed calls after 20 warm-up calls. LEXIC inputs are trimmed to their true length (113 or 815 fixations); the RoBERTa-base row measures one encoder pass at 510 tokens. Size denotes FP32 model weights.}
\label{tab:eff}
\centering
\footnotesize
\renewcommand{\arraystretch}{1.05}
\setlength{\tabcolsep}{2.4pt}
\begin{tabular}{lcccc}
\toprule
Model & Params & MACs/trial & Size & Latency \\
\midrule
LEXIC (mean length) & $41.6$K & $3.1$M & $166$\,KB & $1.4$\,ms \\
LEXIC (worst case) & $41.6$K & $21.9$M & $166$\,KB & $9.6$\,ms \\
AhnCNN (retrained baseline) & $819$K & $9.0$M & $3.3$\,MB & $0.14$\,ms \\
RoBERTa-base (PLM)~\cite{shubi2024finegrained} & $124.6$M & $48.1$G & $499$\,MB & $122$\,ms \\
\bottomrule
\end{tabular}
\end{table}

\begin{figure}[t]
\centering
\resizebox{\columnwidth}{!}{%
\begin{tikzpicture}[
  font=\sffamily\scriptsize,
  pt/.style={circle, fill=blue!65, draw=black!40, inner sep=1.6pt},
  ptg/.style={circle, fill=black!35, draw=black!40, inner sep=1.6pt},
  lbl/.style={font=\sffamily\scriptsize},
]
\def\X#1{{(#1-4.5)*1.842}}
\def\Y#1{{(#1-0.48)*25}}

\fill[green!12] (\X{7.75},\Y{0.56}) rectangle (\X{8.3},\Y{0.63});
\node[lbl, text=green!45!black, align=center] at (\X{8.02},\Y{0.607})
  {PLM\\[-2pt]models};

\draw[dashed, black!45] (\X{4.5},\Y{0.525}) -- (\X{8.3},\Y{0.525});
\node[lbl, text=black!55, anchor=west] at (\X{6.15},\Y{0.532}) {BEyeLSTM (published)};

\draw[->, black!70] (\X{4.5},\Y{0.48}) -- (\X{8.35},\Y{0.48})
  node[right, lbl, black, xshift=-1mm, yshift=-3.2mm] {parameters};
\draw[->, black!70] (\X{4.5},\Y{0.48}) -- (\X{4.5},\Y{0.645});
\node[lbl, rotate=90, anchor=south] at (\X{4.02},\Y{0.562}) {Pooled AUROC};

\foreach \lx/\tk in {5/100K, 6/1M, 7/10M, 8/100M} {
  \draw[black!70] (\X{\lx},\Y{0.48}) -- (\X{\lx},\Y{0.4765});
  \node[lbl, below, yshift=-0.5mm] at (\X{\lx},\Y{0.4765}) {\tk};
}
\foreach \ly in {0.50, 0.55, 0.60} {
  \draw[black!70] (\X{4.5},\Y{\ly}) -- (\X{4.47},\Y{\ly});
  \node[lbl, left, xshift=0.4mm] at (\X{4.47},\Y{\ly}) {\ly};
}

\node[ptg] (ahn) at (\X{5.913},\Y{0.496611696}) {};
\node[lbl, anchor=west, xshift=1.2mm] at (ahn) {AhnCNN};

\node[pt] (le) at (\X{4.611},\Y{0.517434964}) {};
\node[pt] (lefl) at (\X{4.619},\Y{0.542025875}) {};
\draw[blue!55, thick] (le) -- (lefl);
\node[lbl, anchor=west, xshift=1.4mm, text=blue!55!black] at (lefl) {\textbf{LEXIC}};
\node[lbl, anchor=north west, align=left, xshift=1.4mm, yshift=-0.8mm, text=blue!55!black] at (le) {LEXIC without\\[-2pt]lexical features};
\end{tikzpicture}}
\caption{Pooled AUROC versus model size (log scale) on OneStop. Filled markers pool test predictions across the three settings within each seed and fold, then average AUROC over seeds and folds. The dashed line and shaded band are published pooled results for BEyeLSTM and the best PLM-based models~\cite{shubi2025eyebench}. LEXIC's two variants sit at ${\sim}41$K parameters.}
\label{fig:pareto}
\end{figure}
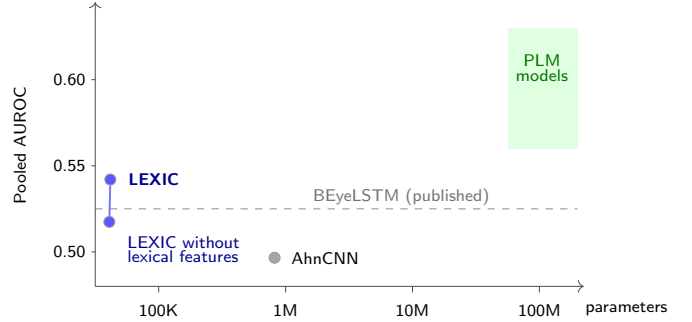

Table~\ref{tab:eff} reports model size and latency. LEXIC uses $166$\,KB of weights and runs in $1.4$\,ms at the mean sequence length. It has about $20$ times fewer parameters than AhnCNN and $3000$ times fewer than the RoBERTa-base reference encoder. AhnCNN runs in $0.14$\,ms, while one RoBERTa-base forward pass takes $122$\,ms under the same measurement protocol. Despite its lower MAC count, LEXIC is slower than AhnCNN in wall-clock latency because recurrent computation is sequential; nevertheless, its 1.4-ms latency remains small for passage-level inference. Fig.~\ref{fig:pareto} compares AUROC and parameter count.

\subsection{Calibration}

For AhnCNN and LEXIC, Platt scaling~\cite{platt1999probabilistic} reduces mean expected calibration error from about $0.27$ to between $0.020$ and $0.052$, using ten bins of equal width. Calibration is fitted on validation data separately for each seed and fold. With thresholds selected on the raw validation scores, LEXIC reaches balanced accuracy of $0.520$/$0.535$ on Unseen Text/Unseen Reader, compared with $0.497$/$0.499$ for \mbox{AhnCNN}. At inference, calibration uses a scalar transform, and classification compares the raw score with a stored threshold.

\section{Discussion}
\label{sec:discussion}

LEXIC combines a compact gaze encoder with frequency and length features. The ablation shows that changing the encoder or adding lexical features to AhnCNN improves prediction on OneStop Unseen Text and Unseen Reader. The complete model provides consistent gains across Unseen Text and Unseen Reader while retaining language-model-free inference.

LEXIC produces one estimate after a reader finishes a passage. The evaluation covers English reading recorded with research eye trackers in OneStop and SB-SAT. The CPU measurements show that the prediction model requires little memory and runs in milliseconds on the tested laptop, indicating its suitability for resource-constrained on-device deployment.

\section{Conclusion}

LEXIC predicts reading comprehension using a compact gaze encoder and two lexical features from lookup tables. It reaches $0.529$/$0.554$ AUROC on OneStop Unseen Text/Unseen Reader, improving over retrained AhnCNN by $3.4$/$5.8$ percentage points. The model uses $166$\,KB of weights and runs in $1.4$\,ms on one laptop CPU core without language model computation.

\bibliographystyle{IEEEtran}
\bibliography{references}

\end{document}